\documentclass[runningheads]{llncs}
\usepackage[T1]{fontenc}
\usepackage{graphicx}
\usepackage{booktabs}
\usepackage[hidelinks]{hyperref}
\usepackage{xurl}
\usepackage{orcidlink}

\renewcommand{\orcidID}[1]{\orcidlink{#1}}

\usepackage{amsmath, amssymb}

\usepackage{algorithm}
\usepackage{algpseudocode}
\begin{document}
\title{A Step Forward Towards Trustworthy Risk-Aware Facial Retrieval (RA-FR)
}
%
%
\author{Muhammad Emmad Siddiqui\inst{1}\orcidID{0009-0005-1575-8409} \and
Muhammad Rafi\inst{1}\orcidID{0000-0002-3673-5979}} 
\authorrunning{M. E. Siddiqui and M. Rafi}
%
\institute{National University of Computer and Emerging Sciences, Karachi, Pakistan}
%
\maketitle              
\begin{abstract}
Facial image retrieval in unconstrained surveillance environments is a high-stakes challenge where missing a subject of interest—a single false negative—is simply not an option. Despite near-perfect performance on curated benchmarks, current recognition systems falter under real-world domain shifts such as low resolution, motion blur, and uncontrolled illumination (e.g., SCFace). Addressing this reliability gap, we propose \textbf{Risk-Aware Facial Retrieval (RA-FR)}, a framework that moves beyond fixed Top-$k$ retrieval to adaptive set generation, guaranteeing ground truth inclusion within a user-specified risk level ($\alpha$) and confidence level ($1-\delta$). Our approach integrates three core contributions: (1) reducing aleatoric uncertainty via a hybrid blind face restoration technique coupling Latent Consistency Models (InterLCM) and DiffBIR; (2) extracting discriminative, restoration-robust features via self-supervised DINOv1 ViT-B with GGeM pooling; and (3) employing conformal prediction with Hoeffding's inequality to dynamically calibrate retrieval set sizes based on query uncertainty. On the IMFDB benchmark, it consistently satisfies a \textbf{5\% risk target} with an average retrieval set size of approximately \textbf{10 images}. By unifying domain-specific restoration, robust representation learning, and provable decision rules, RA-FR offers a pipeline that makes facial retrieval in surveillance both reliable and auditable. The code is available at: \url{https://github.com/MuhammadEmmadSiddiqui/RA-FR}.
\end{abstract}

\section{Introduction}
Facial image retrieval embeds a query into a high-dimensional feature space and returns visually similar database images via nearest-neighbor search (e.g., FAISS)~\cite{FAISS}. Despite widespread deployment, such systems can produce severe failures in high-stakes domains like law enforcement and medicine. Face detection also underpins broader applied monitoring systems, from transit occupancy sensing to access control~\cite{islam2020passenger}. Documented wrongful arrests, including those of Robert Williams and Porcha Woodruff highlight the real-world consequences of unreliable facial matches~\cite{American,CBS2025}. Conventional pipelines select a fixed \(k\) (top-\(k\)) using in-distribution metrics such as recall@\(k\). However, this choice is brittle under distribution shift: a \(k\) tuned on training data provides no deployment-time guarantee that a true match is returned~\cite{Kuhn2025}. While uncertainty estimation methods (e.g., Bayesian Triplet Loss) yield useful confidence signals~\cite{BTL}, a scalar uncertainty score alone does not guarantee retrieval correctness or coverage.

This motivates a risk-aware formulation of retrieval, in which \emph{risk} is formally defined as the probability that the retrieval set fails to include any of the true nearest neighbors of a given query sample. By combining uncertainty estimates with formal risk control, retrieval systems can produce set-valued outputs with explicit coverage guarantees. Unlike fixed-\(k\) retrieval which assumes similarity scores are fully trustworthy, risk-aware retrieval adapts the set size to per-query uncertainty and enforces a user-specified coverage bound \(\alpha\) with confidence \(1-\delta\). When uncertainty is high, the system enlarges the retrieval set to preserve a formal reliability guarantee.

Our RA-FR framework operationalizes this principle using the~\cite{rcir} two-module architecture: an offline \emph{Risk Controller} estimates a global scale parameter \(\kappa\) from a small calibration set, while a lightweight online \emph{Adapter} maps per-query uncertainty to an instance-specific retrieval size \(K_i\). This separation maintains runtime efficiency while providing finite-sample statistical guarantees on retrieval risk. We summarise our main contributions below.
\begin{itemize}
  \item \textbf{Restoration pipeline:} We integrate DiffBIR and InterLCM into the retrieval process to reduce noise and improve calibration.
  \item \textbf{Robust embeddings:} By combining DINOv1-pretrained ViT backbones with GeM/GGeM pooling, we produce sharper, lower-uncertainty descriptors.
  \item \textbf{Validation:} Experiments on the IMFDB~\cite{IMFDB} and SCFace~\cite{SCFace} datasets confirm that RA-FR meets \((\alpha,\delta)\) guarantees and offers better risk--efficiency trade-offs than existing baselines.
\end{itemize}
To our knowledge, this is the first work to unify uncertainty-aware metric learning, formal risk control, DINOv1, and blind face restoration. RA-FR creates a trustworthy framework for high-stakes deployment, trading a small increase in set size for statistically grounded reliability.

\section{Literature Review}
\subsection{Facial Image Retrieval}
Facial retrieval systems have evolved significantly since early implementations such as the CIA's ``Face Trace,'' which utilized geometric cues and auxiliary metadata to generate probabilistic rankings \cite{CIA}. While these foundational systems established the utility of probabilistic outputs, they were often prone to embedding demographic biases within biometric pipelines, a challenge that persists in contemporary surveillance applications \cite{Stevens2021}.
Traditional retrieval methodologies initially relied on handcrafted descriptors, such as SIFT, to encode local keypoints and geometric invariances \cite{Survey}. The field underwent a paradigm shift with the advent of deep Convolutional Neural Networks (CNNs), which demonstrated that learned hierarchical features significantly outperform engineered descriptors. While CNNs provide robust baselines due to large-scale pretraining, their architectural reliance on local receptive fields restricts their ability to model long-range semantic dependencies effectively \cite{Survey}.
\begin{figure}
    \centering
    \includegraphics[width=\linewidth]{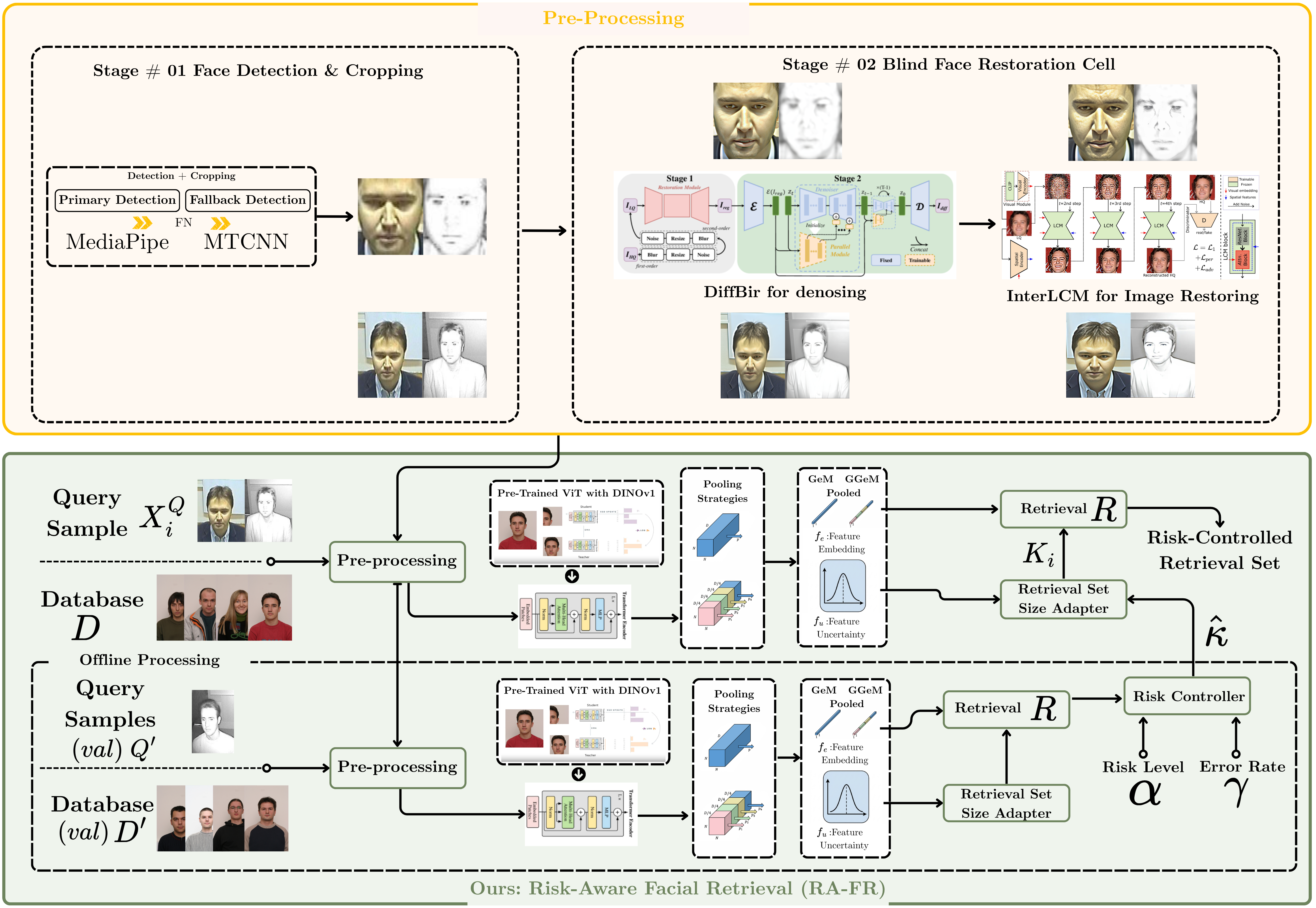}
    \caption{Overall architecture of the proposed Risk-Aware Facial Retrieval framework.}
    \label{fig:ra_fr_pipeline}
\end{figure}

Vision Transformers (ViTs) address these structural limitations by employing global self-attention mechanisms, yielding object-centric embeddings that exhibit greater robustness to occlusion compared to CNNs \cite{ViT_vs_CNNs}. In the facial domain specifically, pyramid ViT designs learn discriminative multi-scale representations that surpass CNNs and plain ViTs with fewer parameters \cite{islam2022fpvt}. Furthermore, self-supervised frameworks such as DINO enable ViTs to learn rich, label-agnostic representations that frequently match or surpass supervised benchmarks \cite{Caron2021}. This integration of global attention with self-supervised learning constitutes the backbone adopted in this work to facilitate robust, uncertainty-aware facial retrieval.
\subsection{Conformal Prediction in Image Retrieval}
\label{sec:cp_background}
The quantification of uncertainty in deep learning is traditionally categorized into \emph{aleatoric} (data noise) and \emph{epistemic} (model ignorance)~\cite{Uncertainties,rcir}. While standard Bayesian approximations~\cite{Dropout,Ensembles} provide uncertainty scores, they lack finite-sample statistical guarantees. 

Standard Conformal Prediction (Split-CP) addresses this by using a hold-out calibration set to rigorously map heuristic scores to prediction sets $\mathcal{C}(X)$~\cite{Intro_to_CP}.  By computing the $(1-\alpha)$-quantile of non-conformity scores (e.g., softmax entropy), Split-CP guarantees marginal coverage: $\Pr(Y \in \mathcal{C}(X)) \ge 1-\alpha$. 

However, standard CP is limited to classification coverage and does not directly apply to complex tasks like retrieval, where the goal is to bound a semantic loss (e.g., $1-\text{Recall}$). To address this, recent works extended CP to \emph{Risk-Controlling Prediction Sets} (RCPS)~\cite{angelopoulos2021risk}. Instead of simple coverage, RCPS seeks a threshold $\lambda$ that bounds a monotonic risk function $R(\lambda)$ below a user-defined tolerance $\alpha$ with probability $1-\delta$.

RCIR~\cite{rcir} adapts the RCPS framework specifically for image retrieval. It employs Hoeffding's inequality to estimate the risk upper bound $\hat{R}^+(\lambda)$ on the calibration set. The system then selects the tightest threshold $\hat{\lambda}$ such that $\hat{R}^+(\hat{\lambda}) \le \alpha$, thereby guaranteeing that the retrieval set contains the true match with high probability. While RCIR establishes the statistical framework, it remains susceptible to high aleatoric noise (e.g., blurred or degraded inputs) while dealing with surveillance images or OOD datasets, which inflates the prediction set sizes to maintain valid coverage. Our work (RA-FR) addresses this limitation by integrating blind restoration and leveraging robust embeddings to tighten the risk bounds without sacrificing statistical validity.
\section{Methodology}
\label{sec:methodology}

Fig.~\ref{fig:ra_fr_pipeline} illustrates our end-to-end Risk-Aware Facial Retrieval (RA-FR) pipeline, spanning from raw input pre-processing to risk-controlled set retrieval. Our design specifically targets surveillance snapshots, which are characterized by severe degradation (blur, noise, low resolution) and are consequently dominated by aleatoric uncertainty~\cite{Uncertainties,face_restoration}. To mitigate these aleatoric factors prior to embedding, we employ a two-stage pre-processing cascade.
\paragraph{Stage I — Face Detection:}
We implement a dual-path detection strategy to balance throughput and robustness. A lightweight detector (MediaPipe) processes the high-speed detection. Failures identified by extreme downsampling, motion blur, or severe illumination anomalies are forwarded to a computationally heavier but more robust MTCNN fallback~\cite{MediaPipe,MTCNN}. Detected faces are cropped and aligned; we simultaneously maintain a \emph{no-crop} ablation stream to empirically quantify the impact of tight facial cropping on downstream retrieval calibration.
\paragraph{Stage II — Blind Face Restoration:}
To recover semantic details from degraded inputs, we employ a restoration chain. The pipeline first applies DiffBIR's~\cite{DiffBIR} Blind Image Denoising V2.1 module, followed by refinement via an InterLCM~\cite{InterLCM} (Latent Consistency Model). To handle the extreme degradation of surveillance imagery, this refinement utilizes a targeted 3-step reconstruction process. Furthermore, InterLCM alignment settings are configured dynamically, applying aligned processing for cropped faces and unaligned processing otherwise. We also evaluate DiffBIR in isolation to determine if the additional computational cost of the LCM refinement yields operational gains in retrieval specifically, whether restoration allows the system to attain a fixed target risk $\alpha$ with a smaller average retrieval set size $K$, rather than merely offering cosmetic improvements.

Post-restoration, images are embedded using a DINOv1-pretrained Vision Transformer backbone~\cite{ViT,Caron2021}. To compact the feature representation, we aggregate patch tokens via Generalized Mean (GeM) and Group Generalized Mean (GGeM) pooling~\cite{GGeM}. The architecture outputs both latent embeddings $f_e$ and predictive uncertainty estimates $f_u$. These signals serve as the input for controlling risk.

To construct valid prediction sets from the raw embeddings, the pipeline executes the RA-FR framework in two distinct phases:
\begin{itemize}
    \item \textbf{Offline Calibration:} We utilize a held-out calibration set (Validation $Q'$ and $D'$) to compute the risk-controlling parameter $\hat{\kappa}$. The \emph{Risk Controller} evaluates non-conformity scores (cosine distances) against the user-defined acceptable risk level $\alpha$ and error rate $\delta$. By inverting the risk function via Hoeffding’s inequality~\cite{angelopoulos2021risk}, we derive the tightest possible threshold $\hat{\kappa}$ that guarantees the risk bound is satisfied with probability $1-\delta$.
    \item \textbf{Online Inference:} During deployment, the \emph{Retrieval Set Size Adapter} dynamically determines the cut-off size $K_i$ for each query $X_i^Q$. Instead of returning a fixed top-$k$ list, the adapter uses the pre-computed $\hat{\kappa}$ to expand or contract the retrieval set based on the query's specific uncertainty. If the query is ambiguous (high uncertainty), the set size $K_i$ increases automatically to maintain the valid risk guarantee; conversely, for clear queries, the set shrinks to maximize efficiency.
\end{itemize}

\subsection{Uncertainty Estimation \& Risk Controlled Mechanism}
\label{subsec:risk_control}

Standard image retrieval uses a feature extractor $fe$ to map images into a space where similarity is determined by distance $d$. For any query image $X_i^Q$, the system retrieves the top-$K$ closest matches from the database $D$, defined as:
\begin{equation}
R(X_i^Q) = \left\{ x \in D \mid \text{rank}(d(fe(X_i^Q), fe(x))) \leq K \right\}
\end{equation}
However, this standard approach lacks a reliability metric. To address this, we employ two complementary methods for uncertainty estimation:

\vspace{2mm}
\noindent \textbf{(1) Monte Carlo Dropout (MCD).} We approximate Bayesian posteriors by performing $T$ stochastic forward passes at test time~\cite{Dropout}. The predictive mean $\mu$ and epistemic uncertainty $\sigma^2$ are computed as:
\begin{equation}
    \mu = \frac{1}{T} \sum_{t=1}^{T} f_{e}^{(t)}(X), \quad \sigma^2 = \frac{1}{T} \sum_{t=1}^{T} \left( f_{e}^{(t)}(X) - \mu \right)^2
\end{equation}

\vspace{2mm}
\noindent \textbf{(2) Single Deterministic Model.} To reduce computational cost, we model features as Gaussian distributions $\mathcal{N}(\mu, \sigma^2)$ using a dedicated variance head $f_u$ trained via Bayesian triplet loss~\cite{BTL}. This yields uncertainty in a single pass:
\begin{equation}
    [\mu, \sigma^2] = [f_e(X), f_u(X)]
\end{equation}

We apply both MCD and BTL uncertainty estimation techniques across multiple backbone architectures listed in the Tab. ~\ref{tab:model_specs}. Alongside the ResNet-50 baseline from~\cite{rcir}, we evaluate our proposed ViT-B models incorporating GeM and GGeM pooling strategies. Their resulting Expected Calibration Error (ECE), and overall retrieval performance is detailed in section~\ref{subsection:FUE} and~\ref{Ablation_Study}.

\begin{table}[ht]
\centering
\renewcommand{\arraystretch}{1.3} 
\setlength{\tabcolsep}{12pt} 

\caption{Model Architectures and Specifications.}
\label{tab:model_specs}
\begin{tabular}{lccc} 
\toprule
\textbf{Model} & \textbf{Backbone} & \textbf{Pre-training} & \textbf{Pooling} \\
\midrule
ViTB-DN-GGeM & ViT-B (DINOv1) & ImageNet-1k (SSL) & GGeM \\
ViTB-DN-GeM  & ViT-B (DINOv1) & ImageNet-1k (SSL) & GeM \\
R50-GeM      & ResNet50       & ImageNet-1k (Sup.)      & GeM \\
\bottomrule
\end{tabular}
\end{table}

\subsubsection*{Risk-Controlled Adapter \& Controller} The aforementioned methods produce uncertainty scores, but they do not provide statistical guarantees. To obtain provable coverage properties, we adopt the distribution-free risk control~\cite{angelopoulos2021risk,rcir}.

\textbf{Problem Formulation.} Let $\mathcal{D}_{\text{cal}} = \{(X_i, Y_i)\}_{i=1}^n$ be an i.i.d. held-out calibration set. For a given query $X$, the \textit{RA-FR Adapter} constructs a dynamic retrieval set $\mathcal{T}_\kappa(X)$ of size $K(X, \kappa)$, scaled by a global parameter $\kappa \in \mathbb{R}^+$:

\begin{equation}
    \label{eq:adapter}
    K(X, \kappa) = \lceil \kappa \cdot \Phi[f_u(X)] \rceil.
\end{equation}

We define the loss as the miscoverage indicator function $\ell(\kappa, X, Y) = \mathbb{I}(Y \notin \mathcal{T}_\kappa(X))$. The population risk ${\rho}(\kappa) = \mathbb{E}[\ell(\kappa, X, Y)]$ constitutes a monotonically non-increasing function with respect to $\kappa$; as the expanded set size $K$ increases, the probability of missing the true match strictly decreases.

\begin{algorithm}
\caption{RA-FR Controller Calibration}\label{alg:calibration}
\begin{algorithmic}[1]
\State \textbf{Input:} Calibration set $\mathcal{D}_{\text{cal}} = \{(X_i, Y_i)\}_{i=1}^n$, Target risk $\alpha$, Error rate $\delta$.
\State \textbf{Output:} Calibrated parameter $\hat{\kappa}$.
\State Compute uncertainty scores $u_i = \Phi(f_u(X_i))$ for all $i$.
\State Define a grid of candidate $\kappa$ values: $\mathcal{K} = \{k_1, \dots, k_M\}$.
\For{$\kappa \in \mathcal{K}$}
    \State Compute set sizes $K_i = \lceil \kappa \cdot u_i \rceil$.
    \State Calculate empirical risk $\hat{\rho}(\kappa) = \frac{1}{n} \sum_{i=1}^{n} \mathbb{I}(Y_i \notin \text{Top-}K_i)$.
    \State Compute bound $\hat{\rho}^+(\kappa) = \hat{\rho}(\kappa) + \sqrt{\frac{\ln(1/\delta)}{2n}}$.
\EndFor
\State $\hat{\kappa} \leftarrow \min \{ \kappa \in \mathcal{K} : \hat{\rho}^+(\kappa) \leq \alpha \}$.
\State \textbf{Return} $\hat{\kappa}$.
\end{algorithmic}
\end{algorithm}

\textbf{Calibration via Risk Control.} The \textit{RA-FR Controller} seeks the optimal parameter $\hat{\kappa}$ such that the expected loss is bounded by a user-specified risk tolerance $\alpha \in (0,1)$ with probability at least $1-\delta$. We evaluate the empirical risk on $\mathcal{D}_{\text{cal}}$:
\begin{equation}
    \label{risk}
    \hat{\rho}(\kappa) = \frac{1}{n} \sum_{i=1}^n \ell(\kappa, X_i, Y_i).
\end{equation}
To account for finite-sample variability, we compute the Upper Confidence Bound (UCB), denoted as $\hat{\rho}^+(\kappa)$, via Hoeffding's inequality:
\begin{equation}
    \hat{\rho}^+(\kappa) = \hat{\rho}(\kappa) + \sqrt{\frac{\ln(1/\delta)}{2n}}.
\end{equation}
Leveraging the monotonicity of the loss, we can efficiently determine the tightest valid parameter by searching over a discrete grid of candidate values $\mathcal{K}$:
\begin{equation}
    \hat{\kappa} = \inf \left\{ \kappa \in \mathcal{K} : \hat{\rho}^+(\kappa) \le \alpha \right\}. \label{eq:hoeffding_inf}
\end{equation}

This offline calibration routine is formalized in Algorithm~\ref{alg:calibration}. By deploying $\mathcal{T}_{\hat{\kappa}}(X)$ during inference, the system guarantees ${\rho}(\hat{\kappa}) \le \alpha$ with probability $1-\delta$, effectively replacing fixed-size heuristics with provably safe retrieval bounds.

\subsection{Backbone Architecture and Pooling Strategy}
\label{sec:backbone_pooling}

The Risk-Controlled Image Retrieval (RCIR) framework traditionally employs ResNet-based backbones~\cite{ResNet}. However, convolutional architectures prioritize local receptive fields, which can limit the capture of the holistic semantic structures required for stable uncertainty estimation~\cite{CNNs}. In contrast, Vision Transformers (ViTs) utilize global self-attention to model long-range dependencies~\cite{ViT}. We leverage the DINOv1 framework, specifically fine-tuning the backbone on the IMFDB dataset to learn robust, identity-centric representations~\cite{Caron2021}. By utilizing a momentum teacher–student mechanism with aggressive augmentation, DINO encourages view-consistency. This fine-tuning stage is critical: it adapts the self-supervised robustness (originally learned on general objects) to the specific occlusions and illumination shifts inherent in the unconstrained surveillance facial domain. To evaluate the impact of spatial structure on retrieval, we discard the standard \texttt{[CLS]} token and implement two distinct pooling heads on the output patch tokens:

\textbf{1. Generalized Mean Pooling (ViTB-DN-GeM).}
This variant aggregates all $N$ patch tokens into a single global vector. For a set of token activations $\{x_i\}$, the global descriptor is computed as: $\mathbf{v}_{\text{global}} = \left( \frac{1}{N} \sum_{i=1}^{N} x_i^{p} \right)^{1/p}$ where $p$ is a learnable parameter. By pooling globally, this model achieves maximum translation invariance but discards the spatial layout of facial features (e.g., the relative geometry between eyes and mouth), which is critical for fine-grained identity verification.

\textbf{2. Group Generalized Mean Pooling (ViTB-DN-GGeM).}
To recover spatial distinctiveness, we adopt Group Generalized Mean Pooling~\cite{GGeM}. We spatially partition the patch tokens into a fixed $G \times G$ grid (corresponding to distinct regions of the face) and compute a local GeM vector for each cell. The final embedding is the concatenation of these regional descriptors: $\mathbf{v}_{\text{group}} = \operatorname{concat}(\mathrm{GeM}(\mathcal{G}_1), \dots, \mathrm{GeM}(\mathcal{G}_G))$ Since our pre-processing pipeline ensures strict facial alignment (Section~\ref{sec:methodology}), these token groups correspond to consistent semantic regions (e.g., forehead group, chin group) across all images. We hypothesize that GGeM will outperform standard GeM by encoding this structural information, a claim we empirically validate in the Ablation Study (Section~\ref{Ablation_Study}).
\subsection{Blind Face Restoration (BFR) Pipeline}
\label{sec:restoration}
Surveillance imagery typically suffers from severe resolution degradation (inter-ocular distance $<30$ px) and complex, unknown atmospheric distortions. We model this degradation as $y = (x \ast k)\downarrow_{s} + n$, where the blur kernel $k$ and noise $n$ are \textbf{unknown to the system}~\cite{face_restoration}. This necessitates a \textbf{Blind Face Restoration (BFR)} approach~\cite{BFR}, in contrast to non-blind methods that require explicit knowledge of degradation parameters. As illustrated in Fig.~\ref{fig:ra_fr_pipeline}, our pipeline adopts a decoupled BFR strategy:
\begin{enumerate}
  \item \textbf{Stage I (Blind Denoising):} We utilize the \textbf{DiffBIR (v2.1)} blind image restoration module. This stage implicitly estimates the degradation priors to remove heavy sensor noise and blur without needing a reference kernel, yielding a coarse, denoised latent representation $z_{\mathrm{latent}}$~\cite{DiffBIR}.
  
  \item \textbf{Stage II (Semantic Refinement):} We employ the \textbf{InterLCM} (Latent Consistency Model) to recover high-fidelity facial details. By treating the Stage I output as a semantic guide, we perform blind refinement via 3-step consistency. This process generates missing high-frequency details (hallucination) while strictly adhering to the geometric and identity constraints.
\end{enumerate}
\textbf{Theoretical Justification:} By initializing the generative process at an intermediate restored state rather than pure Gaussian noise, we significantly constrain the solution space~\cite{meng2021sdedit}. The model is forced to adhere to the structural guidance provided by Stage I, effectively acting as a "semantic upsampler" rather than an unconstrained image generator. As noted in recent diffusion-based restoration studies~\cite{yue2022difface,zhou2022codeformer}, this strategy mitigates the identity-drift risks inherent in stochastic generative models. Consequently, it preserves biometric fidelity without requiring computationally expensive fine-tuning.

\section{Results and Discussions}
\label{sec:results_discussion}

\subsection{Datasets}
We benchmark RA-FR on IMFDB~\cite{IMFDB} and SCFace~\cite{SCFace} to evaluate robustness in unconstrained and surveillance settings, respectively.

\textbf{Indian Movie Face Database (IMFDB):} IMFDB~\cite{IMFDB} contains 34,512 images of 100 identities from 103 films. Unlike web-harvested benchmarks (e.g., LFW~\cite{LFW}), IMFDB offers dense intra-class sampling with high \textbf{aleatoric uncertainty} (inherent data noise). This complexity is crucial for training models to distinguish between data noise and epistemic ignorance.

\begin{figure}[h]
  \centering
  \includegraphics[width=\linewidth]{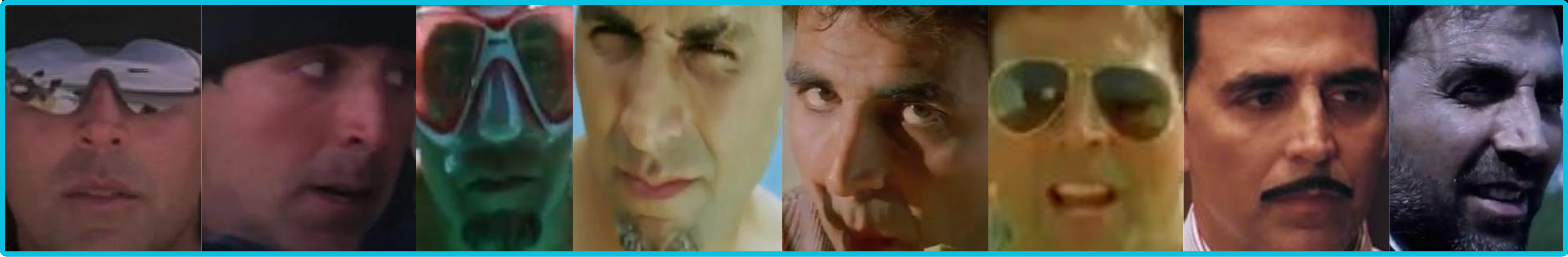}
  \caption{IMFDB samples demonstrating intra-class variability for a single subject.}
  \label{fig:actors-variability}
\end{figure}

Fig.~\ref{fig:actors-variability} illustrates these challenges on a single identity. The samples highlight severe ocular occlusion (sunglasses, masks), extreme lighting contrasts, expression deformations (wincing, shouting), and significant temporal changes (facial hair, aging) and significant temporal changes (facial hair, aging), a factor known to degrade recognition across wide age gaps \cite{islam2021shallow}. RA-FR leverages this diversity to learn robust features, allowing it to adapt retrieval set sizes dynamically even under such high-entropy conditions.

\textbf{SCFace:} To approximate operational surveillance degradations (e.g., blur, compression, IR), we utilise SCFace~\cite{SCFace} as a held-out test set. It comprises 4,160 images of 130 subjects, pairing high-resolution mugshots with low-resolution probes captured at distances of 4.20\,m, 2.60\,m, and 1.00\,m. 

\begin{figure}[h]
  \centering
  \includegraphics[width=\linewidth]{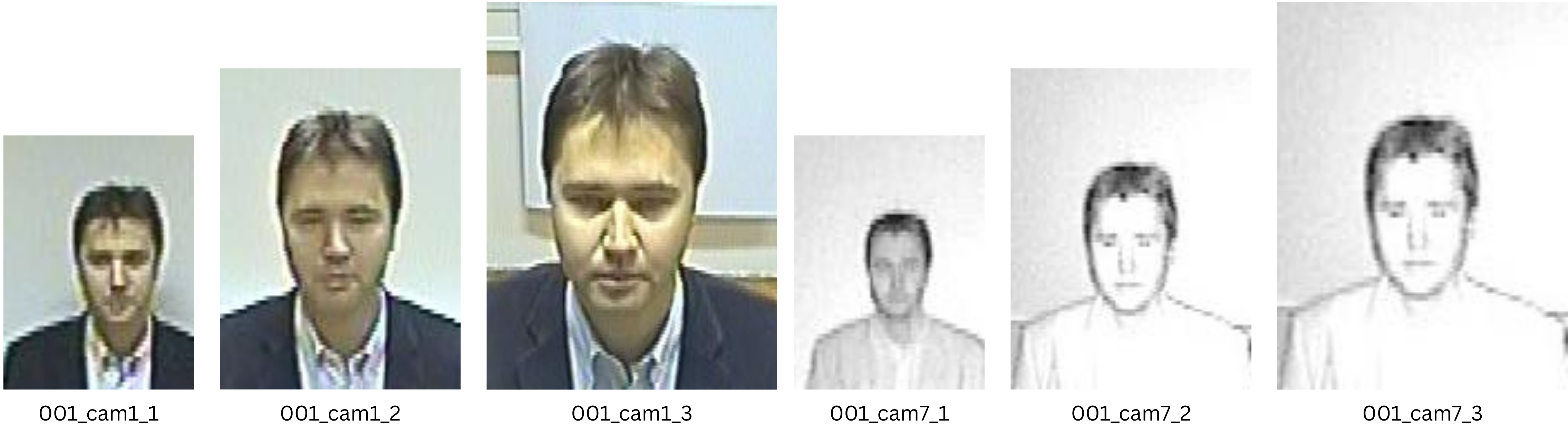}
  \caption{SCFace surveillance samples showing cross-camera and resolution variability.}
  \label{fig:SCFace_Sample_Test}
\end{figure}

Unlike synthetic benchmarks, SCFace introduces realistic \textbf{optics-induced degradations} and camera-dependent noise. This creates a rigorous testbed for cross-resolution matching, evaluating if our risk-controlled retrieval can generalise from high-quality training data (IMFDB) to high-uncertainty, real-world surveillance queries.

\subsection{Feature Representations}
Embeddings are central to risk-aware retrieval; discriminative features directly minimize empirical risk (Eq.~\ref{risk}). We fine-tune DINOv1 (ViT-B), a self-supervised ViT pre-trained on the DINO-ImageNet dataset, and compare two pooling strategies, Generalized Mean (GeM) and Grouped GeM (GGeM), against a standard ResNet50 baseline pre-trained on ImageNet used in~\cite{rcir}.



\begin{figure}[h]
  \centering
  \includegraphics[width=\linewidth]{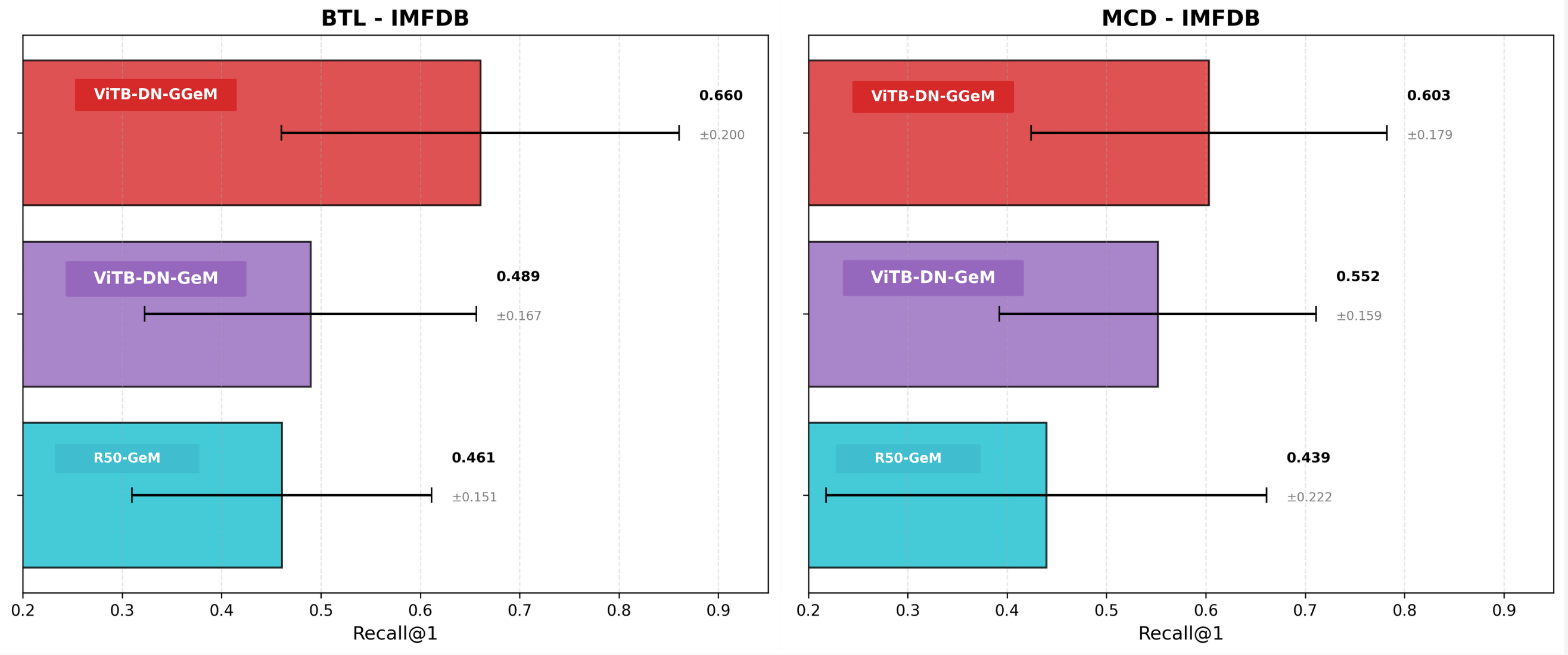}
  \caption{Recall@1 performance comparing MCD and BTL techniques on IMFDB.}
  \label{fig:Recall_Comparison}
\end{figure}

We evaluate performance using Bayesian Triplet Loss (BTL) and Monte Carlo Dropout (MCD). As shown in Fig.~\ref{fig:Recall_Comparison}, ViTB-DN-GGeM achieves $\approx$40\% higher recall than R50-GeM across both uncertainty strategies~\cite{rcir}. This performance leap highlights the superiority of self-supervised ViTs in extracting robust features from unconstrained faces.

\begin{figure}[h]
  \centering
  \includegraphics[width=\linewidth]{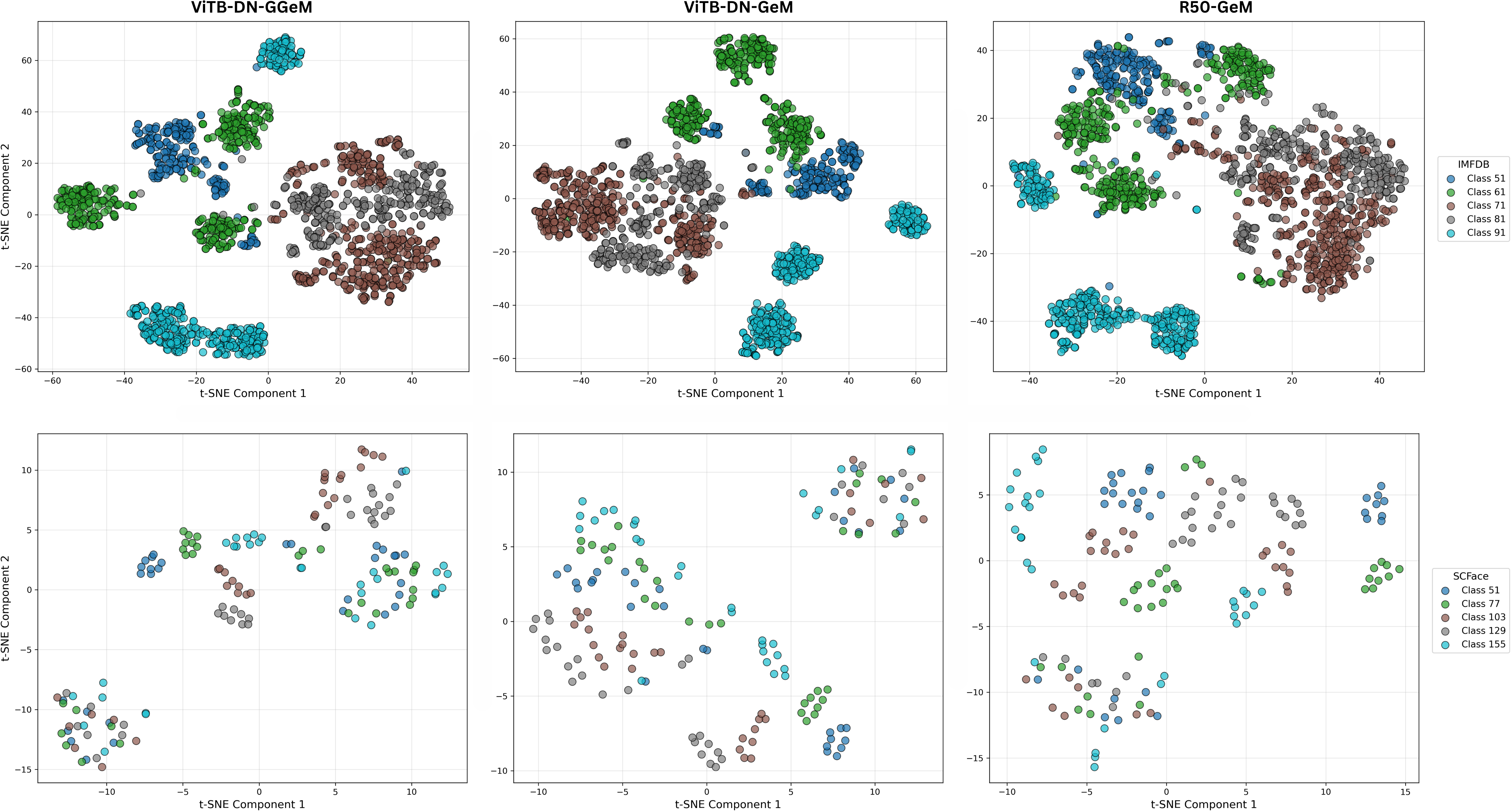}
  \caption{t-SNE projections of embeddings on IMFDB and SCFace datasets.}
  \label{fig:Actors_SCFace_tsne}
\end{figure}

To assess robustness, we visualize t-SNE projections of embeddings from models trained on IMFDB. Fig.~\ref{fig:Actors_SCFace_tsne} (top) shows that under matched-distribution conditions (IMFDB test set), all models produce compact clusters, though ViT variants exhibit tighter intra-class cohesion.

However, under cross-domain evaluation on SCFace (Fig.~\ref{fig:Actors_SCFace_tsne}, bottom), the distribution shift is evident. While cluster compactness degrades for all models, ViTB-DN variants maintain significantly better separation than R50-GeM. The ResNet baseline shows substantial class overlap (low separability), which directly correlates to higher false positive rates and the larger retrieval set sizes needed to satisfy risk guarantees.

\subsection{Feature Uncertainty Estimator}
\label{subsection:FUE}
The feature uncertainty estimator $f_u$ is the engine of our risk-control mechanism. Since the global controller scale $\kappa$ is fixed after calibration (Eq.~\ref{eq:adapter}), the dynamic sizing of retrieval sets depends entirely on the quality of $f_u$. Accurate uncertainty estimates are therefore crucial; a well-calibrated estimator yields tighter set boundaries and smaller retrieval sizes for a given risk threshold.

\begin{figure}[h]
  \centering
  \includegraphics[width=\linewidth]{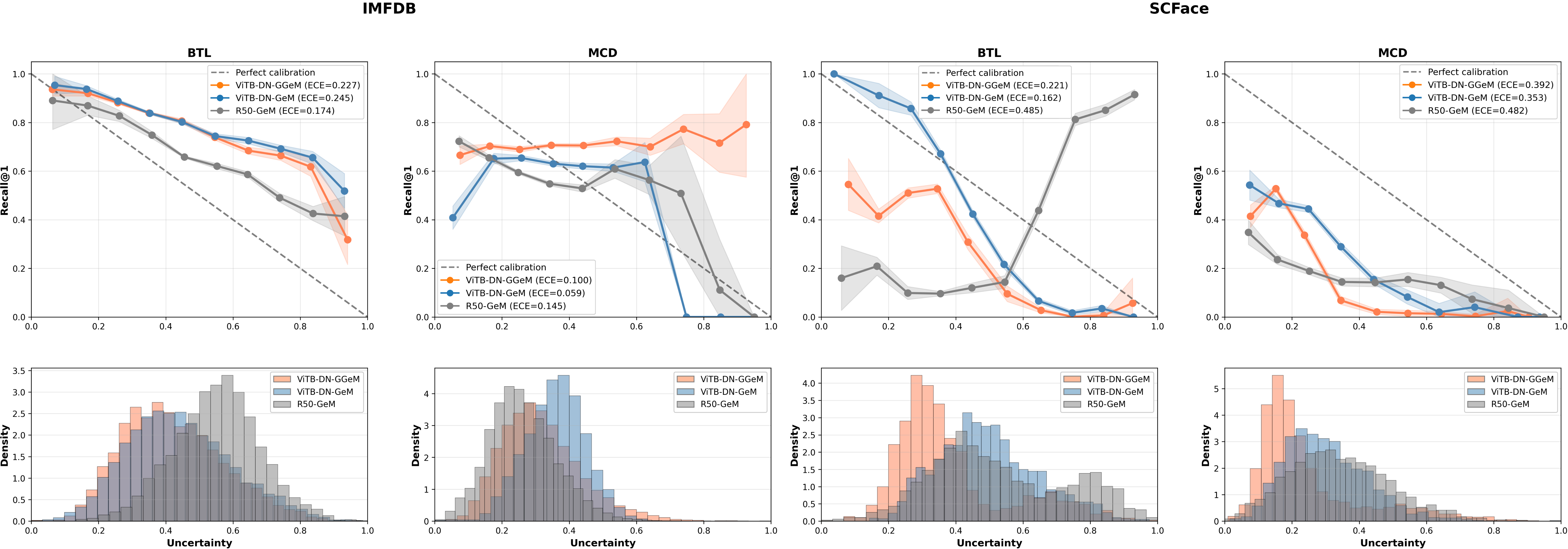}
  \caption{Model calibration (top) \& uncertainty densities (bottom) on IMFDB \& SCFace.}
  \label{fig:ECE_IMFDB_SCFace}
\end{figure}

We measure this quality using \textbf{Expected Calibration Error (ECE)} on IMFDB and SCFace, where an ECE near zero indicates that the predicted uncertainty accurately reflects the probability of retrieval failure. In Fig.~\ref{fig:ECE_IMFDB_SCFace}, the dashed diagonal represents ideal calibration, while the shaded regions denote the standard deviation across 10 trials.

Fig.~\ref{fig:ECE_IMFDB_SCFace} (left) details In-Distribution (IMFDB) performance. Under Bayesian Triplet Loss (BTL), the R50-GeM baseline achieves the lowest ECE, whereas under Monte Carlo Dropout (MCD), the ViTB-DN variants demonstrate superior calibration. 

Crucially, Fig.~\ref{fig:ECE_IMFDB_SCFace} (right) reveals the impact of domain shift. When Same models trained on IMFDB tested on SCFace, the ECE for R50-GeM rises sharply, confirming that estimators tuned on clean data often become miscalibrated (overconfident) on OOD inputs. **Our proposed pipeline significantly mitigates this breakdown:** by pre-emptively restoring inputs and leveraging robust ViT features, it roughly halves the OOD ECE. This improves thr risk control calibration process is what enables our risk controller to function more effectively and reliably even under severe surveillance-style degradation.

\subsection{Risk Control Effectiveness}
\begin{figure}
  \centering
  \includegraphics[width=\linewidth]{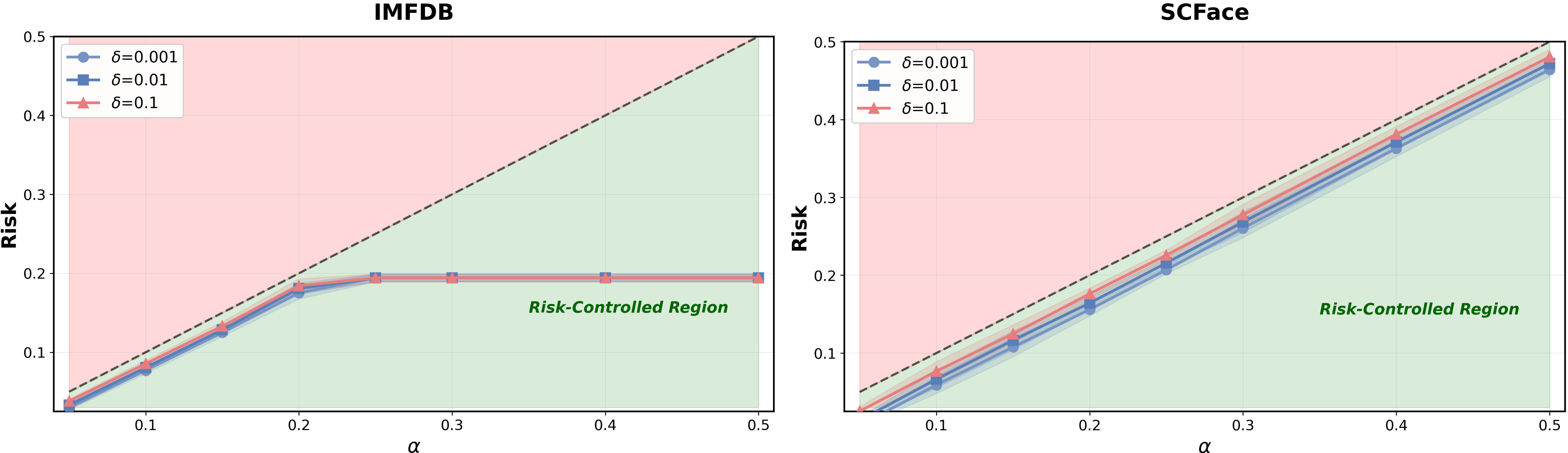}
  \caption{Empirical vs. target risk ($\alpha$) on IMFDB and SCFace across varying $\delta$.} 
  \label{fig:Risk}
\end{figure}

We evaluate the RA-FR procedure across a spectrum of pre-specified target risk levels on both IMFDB and SCFace. Fig.~\ref{fig:Risk} presents the empirical outcomes: the green region below the dotted diagonal indicates instances where the empirical risk successfully remains below the user-defined target $\alpha$, while the red region denotes violations where risk exceeds $\alpha$.

The parameter $\delta$ governs the conservativeness of the calibration. Lower values of $\delta$ impose a stricter confidence requirement (i.e., a larger statistical safety margin). In practice, this forces the adapter to produce larger retrieval sets to guarantee risk satisfaction, thereby prioritizing safety over efficiency. This dynamic is clearly visible in Fig.~\ref{fig:Risk}: as $\delta$ decreases, the empirical curves shift deeper into the green region across both datasets, indicating that the system is "over-satisfying" the risk constraint to ensure validity with high probability.
  
\subsection{Ablation Study}
\label{Ablation_Study}

\begin{figure}
  \centering
  \includegraphics[width=\linewidth]{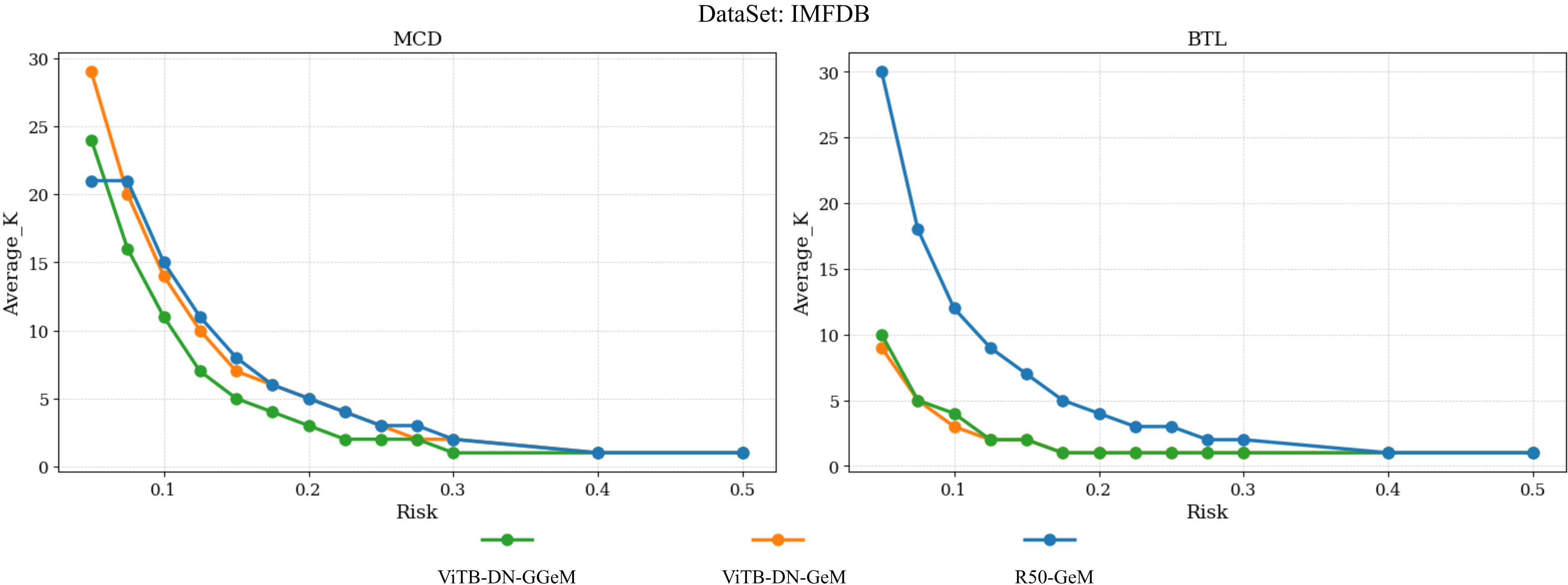}
  \caption{Average retrieval set size ($K$) vs. target risk ($\alpha$) on the IMFDB dataset.}
  \label{fig:Average k without pre-processing}
\end{figure}

Richer embeddings improve not only retrieval recall but also the efficiency of risk-controlled retrieval. Fig.~\ref{fig:Average k without pre-processing} compares R50-GeM, ViTB-DN-GeM, and ViTB-DN-GGeM on the IMFDB Actors split using two uncertainty estimators: Monte Carlo Dropout (MCD) and Bayesian Triplet Loss (BTL). We plot $\mathrm{Average}_K$ (the mean retrieval-set size) against the target risk $\alpha$. Across both estimators, the ViT-based architectures achieve the same empirical risk while requiring substantially smaller $\mathrm{Average}_K$, indicating that semantically richer features allow the risk controller to be more selective.

The effect is most pronounced under BTL at low risk: at $\alpha = 0.05$, R50-GeM retrieves approximately $30$ images on average, whereas ViTB-DN-GeM and ViTB-DN-GGeM retrieve approximately $10$ or fewer, a threefold reduction in retrieval cost. The gap decreases as $\alpha$ increases, showing that backbone superiority is most critical when strict risk guarantees are required.

Upgrading the embedding backbone reduces epistemic uncertainty and shrinks retrieval sets, but this alone is insufficient when queries exhibit large aleatoric uncertainty (e.g., heavy noise, blur, or extremely low resolution). To address input noise, we apply blind face restoration (BFR) to recover identity cues exploitable by downstream recognizers. We evaluate three preprocessing strategies: \emph{No Preprocessing}, \emph{BiD} (DiffBIR-based denoising), and \emph{BiD+InterLCM} (DiffBIR denoising followed by InterLCM latent-consistency refinement).

\begin{figure}
  \centering
  \includegraphics[width=\linewidth]{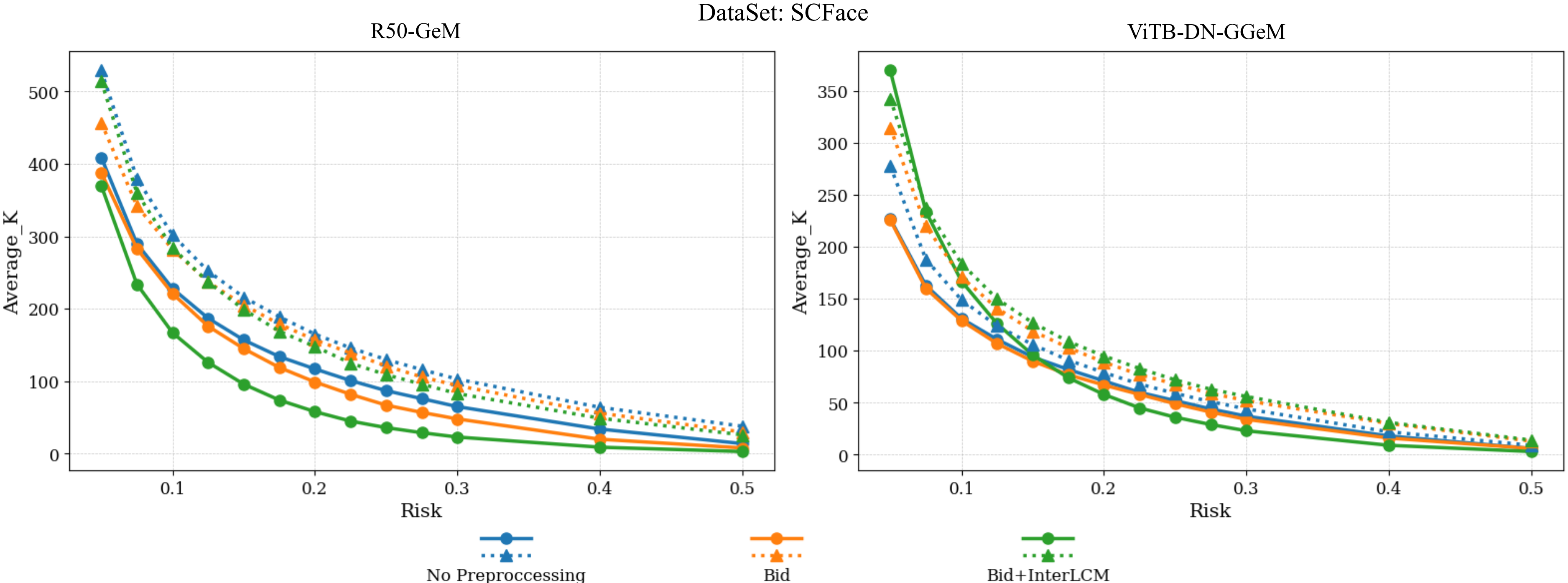}
  \caption{Average $K$ vs. risk ($\alpha$) on SCFace with and without pre-processing steps.}
  \label{fig:Average k with pre-processing}
\end{figure}

Fig.~\ref{fig:Average k with pre-processing} summarizes this trade-off on the SCFace testbed (OOD), with models trained on IMFDB. We report results for two embedding architectures (R50-GeM and ViTB-DN-GGeM), three preprocessing strategies, and two cropping modes (tight face cropping versus full-frame input; cropped queries are shown as dotted curves).

Overall, restoration reduces $\mathrm{Average}_K$ across most risk levels compared to no preprocessing, confirming that targeted enhancement is beneficial under domain shift. However, aggressive cropping can be detrimental for extremely low-resolution queries. SCFace contains faces as small as $28 \times 28$ pixels; tight cropping effectively removes contextual and structural cues required by the enhancement modules. For example, at $\alpha = 0.05$, \textbf{BiD} without cropping retrieves approximately $226$ images on average with ViTB-DN-GGeM, compared to approximately $388$ images with R50-GeM, both under BTL uncertainty estimation. This implies that the ResNet baseline requires retrieval sets \textbf{roughly 70\% larger} to satisfy the same safety constraint. Without preprocessing, the same risk level requires $227$ images with ViTB-DN-GGeM and $409$ with R50-GeM, confirming that embedding quality remains a dominant factor even when restoration is applied.

In summary, (i) stronger embeddings reduce epistemic uncertainty and permit smaller candidate sets; (ii) image restoration helps to further lower $\mathrm{Average}_K$ under domain shift; and (iii) aggressive cropping of extremely low-resolution faces can harm performance. 

\section{Conclusion}

RA-FR introduces a principled framework for uncertainty-aware facial 
retrieval, replacing fixed-size heuristics with a risk-controlling mechanism 
that provably bounds retrieval failure rates. By coupling hybrid blind face 
restoration (InterLCM + DiffBIR) with DINOv1 embeddings aggregated via 
combined GeM and GGeM pooling, we demonstrate that reducing input 
uncertainty via restoration and improving embedding robustness enables the 
risk adapter to satisfy strict safety guarantees ($\alpha=0.05$, confidence 
$1-\delta$) while yielding retrieval sets up to \textbf{70\%} smaller than 
RCIR baselines. Future research should extend this into a two-stage framework 
with a test-time OOD detector to dynamically adjust $\kappa$ under 
distributional shift, and evaluate on more extreme degradation benchmarks 
such as QMUL-SurvFace. RA-FR confirms that reliability in high-stakes 
biometrics requires not just measuring uncertainty, but actively mitigating 
it through domain-specific restoration and robust representation learning.


%
%
%
%





\bibliographystyle{splncs04}
\bibliography{bibliography}

\end{document}